\documentclass[lettersize,journal]{IEEEtran}
\usepackage{amsmath,amsfonts}
\usepackage{algorithmic}
\usepackage{algorithm}
\usepackage{array}
\usepackage[caption=false,font=normalsize,labelfont=sf,textfont=sf]{subfig}
\usepackage{textcomp}
\usepackage{stfloats}
\usepackage{url}
\usepackage{verbatim}
\usepackage{graphicx}
\usepackage{cite}
\usepackage{multirow}
\usepackage{color}
\usepackage{booktabs}
\usepackage[table]{xcolor}
\usepackage{subfloat}
\usepackage{hyperref}
\hypersetup{hidelinks}
\usepackage[switch]{lineno}
\usepackage{cite}

\begin{document}

\title{AFter: Attention-based Fusion Router for RGBT Tracking}

\author{Andong Lu, Wanyu Wang, Chenglong Li, Jin Tang and Bin Luo
\thanks{
This research is partly supported by the National Natural Science Foundation of China (No. 62376004), and the Natural Science Foundation of Anhui Province (No. 2208085J18).
}
\thanks{
     A. Lu, J. Tang and B. Luo are with Anhui Provincial Key Laboratory of Multimodal Cognitive Computation, School of Computer Science and Technology, Anhui University, Hefei 230601, China. 
    (e-mail: adlu\_ah@foxmail.com; tangjin@ahu.edu.cn; luobin@ahu.edu.cn)
    
    Wanyu Wang, and C. Li is with Information Materials and Intelligent Sensing Laboratory of Anhui Province, Anhui Provincial Key Laboratory of Multimodal Cognitive Computation, School of Artificial Intelligence, Anhui University, Hefei 230601, China. 
    (e-mail: 2878065167@qq.com; lcl1314@foxmail.com)
    
}
}

\markboth{Journal of \LaTeX\ Class Files,~Vol.~14, No.~8, August~2021}%
{Shell \MakeLowercase{\textit{et al.}}: A Sample Article Using IEEEtran.cls for IEEE Journals}


\maketitle

\begin{abstract}
Multi-modal feature fusion as a core investigative component of RGBT tracking emerges numerous fusion studies in recent years. However, existing RGBT tracking methods widely adopt fixed fusion structures to integrate multi-modal feature, which are hard to handle various challenges in dynamic scenarios. To address this problem, this work presents a novel \emph{A}ttention-based \emph{F}usion rou\emph{ter} called AFter, which optimizes the fusion structure to adapt to the dynamic challenging scenarios, for robust RGBT tracking. 
In particular, we design a fusion structure space based on the hierarchical attention network, each attention-based fusion unit corresponding to a fusion operation and a combination of these attention units corresponding to a fusion structure. Through optimizing the combination of attention-based fusion units, we can dynamically select the fusion structure to adapt to various challenging scenarios.
Unlike complex search of different structures in neural architecture search algorithms, we develop a dynamic routing algorithm, which equips each attention-based fusion unit with a router, to predict the combination weights for efficient optimization of the fusion structure.
Extensive experiments on five mainstream RGBT tracking datasets demonstrate the superior performance of the proposed AFter against state-of-the-art RGBT trackers. We release the code in \hyperlink{https://github.com/Alexadlu/AFter}{https://github.com/Alexadlu/AFter}.
\end{abstract}


\section{Introduction}
\IEEEPARstart{R}{GBT} tracking aims to fully leverage the complementary advantages of RGB and thermal information for robust visual tracking, which plays an important role in video surveillance~\cite{jain2024fusion}, pedestrian tracking~\cite{zhang2023multi} and multi-sensor information fusion~\cite{fan2024querytrack}.
In recent years, extensive efforts have been made in feature fusion between RGB and thermal modalities ~\cite{2020CMPP,2021MANet++,cui2022visual,Zhang_CVPR22_VTUAV,DMCNet2022,zhang2023efficient,TBSI}, thus pushing forward the development of RGBT tracking. 
However, due to the insufficient exploration of the fusion structure, there is still a lot of research space.

\begin{figure}[t]
    \centering
    \includegraphics[width=0.95\linewidth]{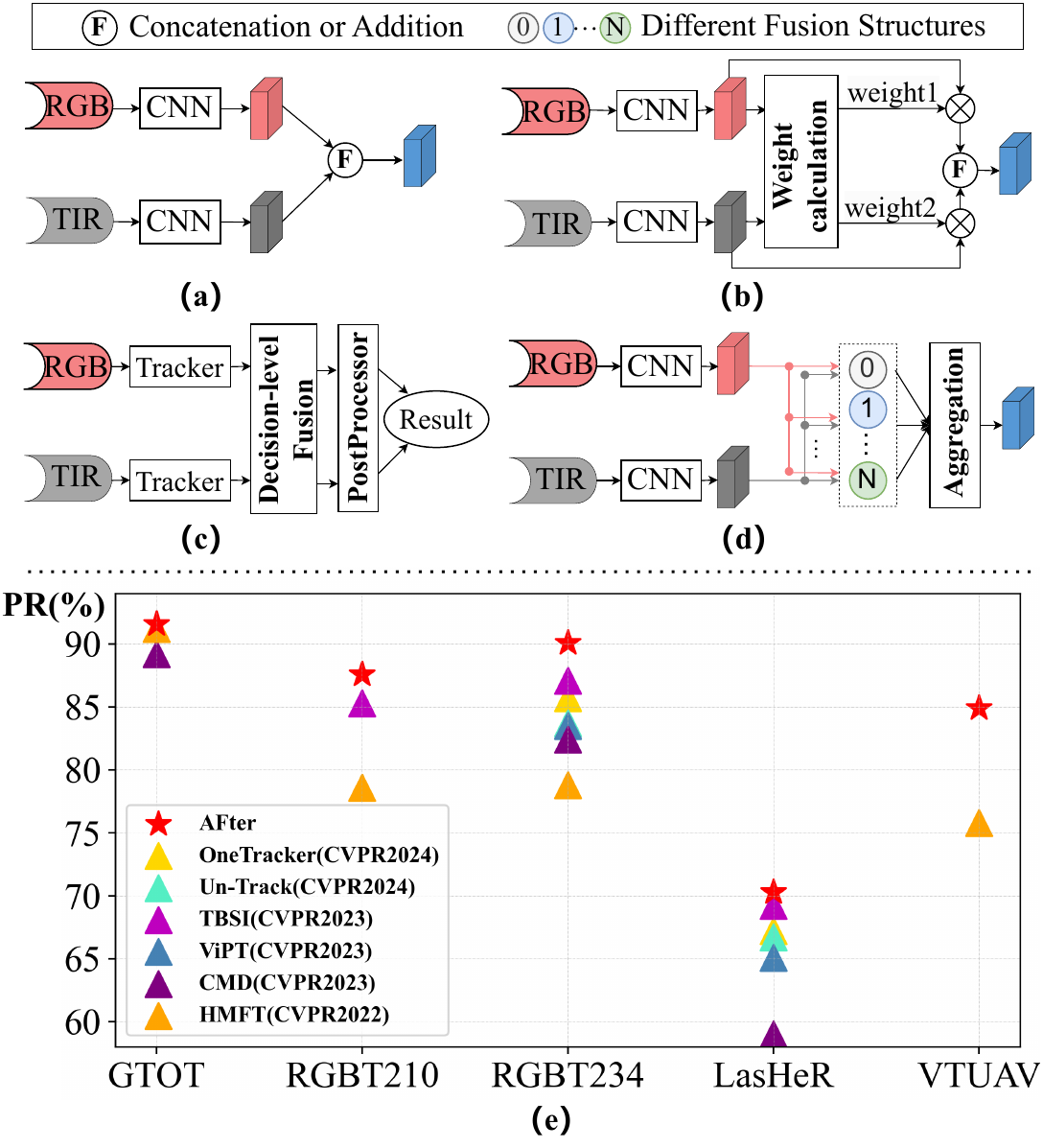}
     \caption{Comparison of existing RGBT tracking models. (a), (b) and (c) indicate representation model, feature fusion model and decision fusion model. (d) denotes the proposed attention-based fusion router. (e) represents the performance comparison between our proposed method and state-of-the-art RGBT tracking methods on precision (PR) scores across five RGBT tracking datasets.}
    \label{motivation_fig}
\end{figure}


Existing RGBT trackers can be roughly divided into two categories. The first category aims to enhance modality representation by designing various representation learning models. For instance, Lu~\emph{et al.}~\cite{2021MANet++} propose a multi-adapter framework that simultaneously extracts modality-shared and modality-specific features. Although these works achieve impressive performance by building robust representational models, they overlook the design of the fusion structure. As depicted in Fig.~\ref{motivation_fig} (a), they fuse two modality features through simple addition or concatenation operations, which restricts the sufficient utilization of modality cooperation information. 
The second category focuses on the fusion structure of RGB and thermal modality features. Most approaches are weighted fusion architectures based on modality importance, as shown in Fig.~\ref{motivation_fig} (b). For example, Liu~\emph{et al.}~\cite{QAT2023} design a quality-aware network to predict the quality of each modality, and convert it into modal feature fusion weights to enhance the role of high-quality modality features in fusion. Cao~\emph{et al.}~\cite{BAT2024}proposes to establish bidirectional adapters between each layer of the backbone network to fully exchange the information of both modalities, thus effectively improving the tracking performance.
In addition, the decision fusion strategy~\cite{tang2023exploring} depicted in Figure \ref{motivation_fig} (c) is also explored in RGBT tracking. 
However, these studies still adopt a fixed fusion structure to deal with all tracking scenarios, which is hard to achieve optimality in different tracking scenarios.

To address this issue, we propose a novel \textbf{A}ttention-based \textbf{F}usion Rou\textbf{ter} (AFter) for RGBT tracking. AFter dynamically adjusts the fusion structure to ensure the optimal fusion for the current input multi-modal features. From the perspective of feature fusion between two modalities, typical feature fusion structures can be divided into three cases: self-fusion, unidirectional fusion and bidirectional fusion. 
For instance, Zhu~\emph{et al.}~\cite{ViPT} propose a channel-wise spatial attention module, which builds a unidirectional fusion from thermal to visible modality.
Hui~\emph{et al.}~\cite{TBSI} designs a template-bridged cross-attention module that allows bidirectional fusion between visible and infrared modalities.
However, these methods typically employ a fixed fusion structure for all scenarios, thus limiting their fusion robustness in tracking scenarios over time.
%
Therefore, we design a dynamic fusion structure space that includes all the above possibilities, allowing different fusion structures to be adopted in different tracking scenarios to improve the overall fusion effect. As present in Figure \ref{motivation_fig} (e), AFter can achieve the performance advantage on all current RGBT tracking datasets. 

In particular, we design a hierarchical attention network (HAN) to provide a dynamic fusion structure space between the two modalities. Each layer of HAN contains four different attention-based fusion units: a spatial enhancement unit, a channel enhancement unit, and two cross-modal enhancement fusion units. In the spatial enhancement unit and the channel enhancement unit, each modality performs intra-modal feature fusion operations from the spatial and channel dimensions, respectively. Two cross-modal enhancement fusion units perform two unidirectional inter-modal feature fusion, respectively. Importantly, when two cross-modal enhancement fusion units are executed simultaneously, a bidirectional inter-modal feature fusion is established. To balance fusion structure diversity and computational efficiency, HAN stacks the four fusion units into three layers in the depth dimension to expand the fusion structure space. In addition, each fusion unit is embedded by a router to predict the combined weight of that unit. Consequently, HAN can construct different fusion structures by combining different fusion units, and dynamically select the fusion structure suitable for the current scene.


To our best knowledge, it is the first time to leverage the dynamic fusion structures in the field of RGBT tracking. Our contributions can be summarized as follows.
\begin{itemize}
   
 \item We propose a novel attention-based fusion router to handle the limitation of fixed fusion structures under dynamic challenging scenarios in RGBT tracking.
 
 \item We design a hierarchical attention network, which provides the space of typical fusion structures to explore the effective fusion scheme for different modalities.
 
 \item We develop a dynamic routing algorithm to efficiently select an effective fusion structure for each frame according to the current scenario.
 
 \item Experiments on five RGBT tracking benchmarks show that the proposed method achieves leading comprehensive performance against state-of-the-art methods.
\end{itemize}

\section{Related Work}
In this section, we will provide a brief overview of the relevant research, focusing on two fields, including RGBT tracking, dynamic networks and attention mechanism. 

\subsection{RGBT Tracking}
In recent years, significant progress has been made in RGBT tracking research, leading to the development of numerous innovative algorithms. Existing studies can be classified into two main categories. The first category focuses on designing different modality representation methods to fully mine each modality information. For instance, Zhang~\emph{et al.}~\cite{ADRNet2021} design a multi-branch network based on challenge attributes to model different challenge scenarios under modalities, thus refining each modality representation. Similarly, Xiao~\emph{et al.}~\cite{APFNet2022} propose to model both modality-shared and modality-specific challenge scenario representations. Shen~\emph{et al.}~\cite{shen2022rgbt} propose a graph learning method based on low-rank decomposition to learn robust RGBT target representations, which jointly solves the low-rank decomposition graph structure and node weights in a unified optimization framework. Despite the impressive progress of these works, neglecting the design of fusion strategies limits the full utilization of modal collaboration information.
The second category is devoted to the design of various clever fusion modules. For example, Liu~\emph{et al.}~\cite{QAT2023} use a modality-based quality evaluator and convert the quality prediction results into modal weights for adaptive fusion. 
Tang~\emph{et al.}~\cite{tang2023exploring} propose an online adaptive decision-level fusion method for extracting the complementary information conveyed by the two modalities. Yuan~\emph{et al.}~\cite{yuan2024improving} perform RoI feature fusion for each pair of aligned RoIs by the Complementary Fusion Transformer (CFT) module, which enhances the modality itself features while capturing the complementary features from another modality.
However, existing fusion methods are typically fixed and struggle with handling diverse complex tracking scenarios simultaneously. 
Unlike these works, this paper innovatively proposes a dynamic multi-strategy fusion method by online adjusting the fusion structure to meet the fusion requirements in different scenarios.

\subsection{Dynamic Routing Methods}
Dynamic networks have become a popular research area in recent years~\cite{hu2023dynamic,li2023dynamask}. 
Unlike static inference neural network structure searches~\cite{lin2017runtime,liu2022toward}, these networks generate real-time execution paths tailored to the input samples. At present, it has been effectively applied to various fields because of its high efficiency. For example, early dynamic approaches focus on network compression via channel pruning~\cite{you2019gate} or layer skipping~\cite{wang2018skipnet}. In addition, a dynamic mechanism is used to solve the problem of scale change. Specifically, the multi-scale features in semantic segmentation \cite{li2020learning} and object detection \cite{song2020fine} are searched using a dynamic routing network. Wu~\emph{et al.}~\cite{wu2018blockdrop} introduce a method that can dynamically select the execution depth during inference to dynamically reduce the total amount of computation. In the field of multimodal analysis, Tsai~\emph{et al.}~\cite{tsai2020multimodal} dynamically adjust the weights between the input modality and the output representation using multimodal routing for each input sample for multimodal language analysis, which identifies the relative importance between features of multiple modalities. Zeng~\emph{et al.}~\cite{zeng2024feature} propose a modal interaction module to learn intra-modal and inter-modal interactions for multimodal sentiment analysis. Despite these advances, the potential of dynamic strategies in RGBT tracking is largely unexploited. 

\subsection{Attention Mechanism} 
Attention mechanisms aim to draw on the human learning process, that is, to facilitate model learning by selecting important information and paying attention to it. In recent years, various attentions are proposed and employed on various visual tasks~\cite{hu2018squeeze, wang2018non, vaswani2017attention, guo2022attention}, and existing attention can be roughly three categories, including channel attention~\cite{hu2018squeeze}, spatial attention~\cite{jia2023multiscale} and self-attention~\cite{wang2018non}. Each kind of attention has a different effect in visual tasks.
Inspired by these attentions, several studies introduce attention strategies in multi-modal fusion to achieve adaptive modal feature interaction. For example, Li~\emph{et al.}~\cite{li2024crossfuse} propose a novel cross attention to enhance the complementary information (uncorrelation) between RGB and thermal modality image. Zhu~\emph{et al.}~\cite{zhu2023skeafn} propose a feature-wised attention fusion module to enhance critical features’ contribution to the multi-modal fusion process in the channel dimension. In addition, there are also some studies applying attention to achieve modality feature fusion in RGBT tracking. For instance, Feng~\emph{et al.}~~\cite{feng2023learning} propose a novel multi-layer attention on the decision level is proposed for robust RGBT tracking. 
Hui~\emph{et al.}~\cite{TBSI} builds a novel template-bridging fusion block based on cross-attention to achieve bidirectional information interaction between visible and infrared modalities. Feng~\emph{et al.}~\cite{feng2024sparse} design a sparse mixed attention aggregation network to achieve information extraction and integration from two modality images. 
However, existing attention schemes often adopt static single-layer architectures, which struggle to dynamically adjust their structures to adapt to changing inputs in complex tracking tasks, thus limiting the dynamic interaction of modality information.

\section{AFter: Attention-based Fusion Router}
\subsection{Overview}
\begin{figure*}[ht]
    \centering
    \includegraphics[width=1\textwidth]{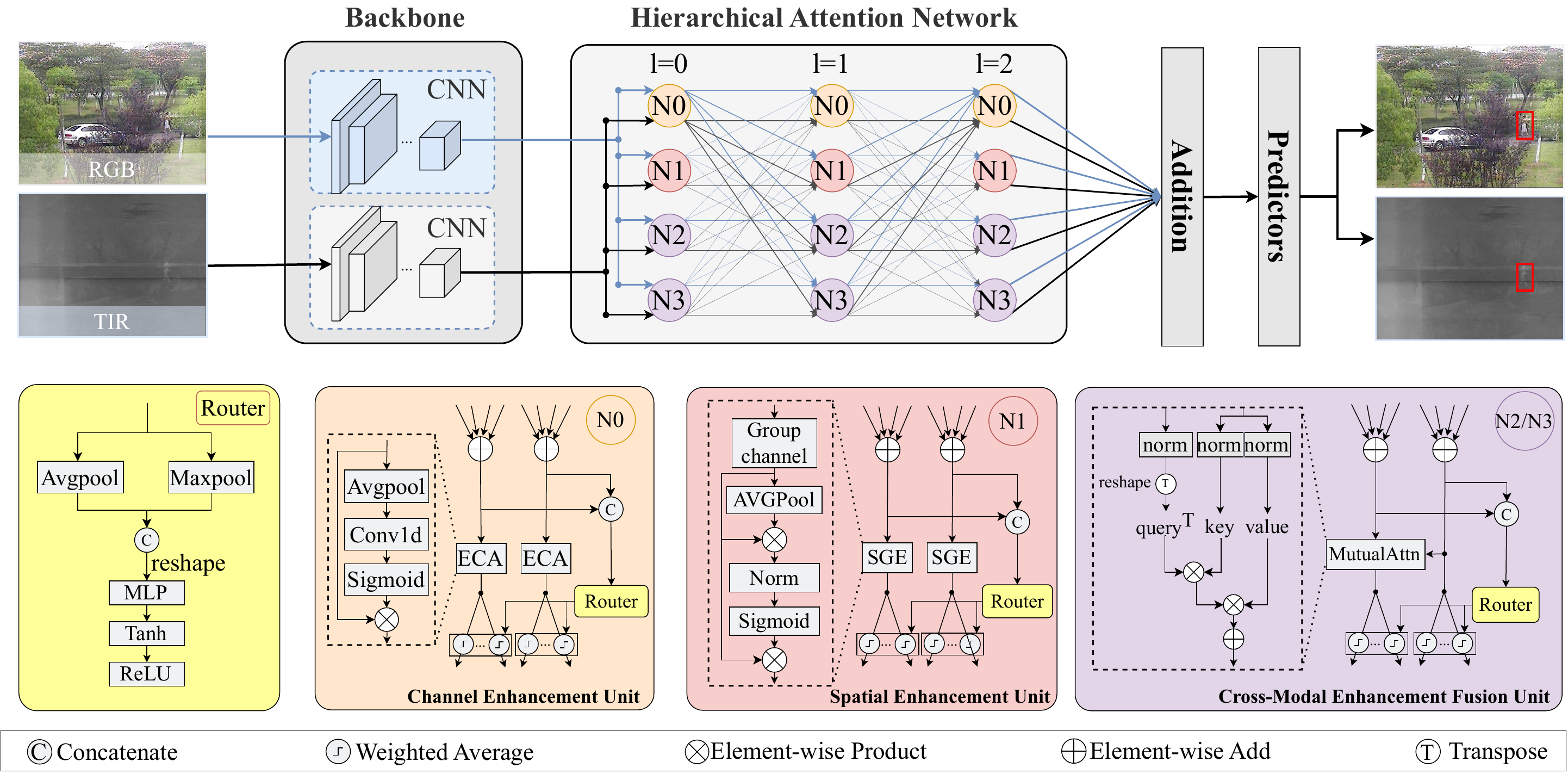}
    \caption{ Overall network architecture of AFter. Illustration of each fusion unit and router. ECA and SGE denote a channel attention~\cite{wang2020eca} and a spatial attention~\cite{li2019spatial} }
    \label{network_overview}
\end{figure*}

In this section, we present the details of the attention-based fusion router (AFter). AFter is a ToMP-based \cite{tomp} RGBT tracker. Firstly, a two-stream feature extraction network is constructed by ResNet-50 \cite{he2016deep} to extract features from visible and thermal modality images respectively. Then, the features of the two modalities are fed into the proposed Hierarchical Attention Network (HAN) for dynamic fusion and output fused features. HAN consists of three layers in total, where each layer contains four different fusion attention units. In addition, each unit is embedded in a router. The detailed structure is shown in Figure \ref{network_overview}. Finally, the fused features are input into the model and IOU predictor for accurate target location and scale prediction.


\subsection{Hierarchical Attention Network}
To cover typical fusion structures between the two modalities, four different fusion units are tailored based on classical attention \cite{li2019spatial,han2018visual,wu2019dual,zhao2020exploring}. These units include the spatial enhancement unit (SEU) and the channel enhancement unit (CEU), as well as the cross-modal enhancement fusion unit for RGB to thermal (CMEU$_{r2t}$) and the cross-modal enhancement fusion unit for thermal to RGB (CMEU$_{t2r}$). The first two units aim to enhance the discriminative cues within each modality, while the latter two units are dedicated to extract cross-modal collaborative information.
In addition, these units are stacked in multiple layers to expand the fusion structure space, where the number of layers is set to 3. The operation of these units can be formally summarized as follows:
\renewcommand{\arraystretch}{1.5}
\begin{equation}
\mathit{O}_i =
\left\{\hspace{-0.3cm}
\begin{array}{lll}
&\mathcal{N}^{(l)}_i(\mathit{F}^{(l)}_i), & i=0 \text{ or } 1, \mathit{F}= rgb \text{ or } tir \\ 
&\mathcal{N}^{(l)}_i(rgb^{(l)}_i, tir^{(l)}_i), & i=2, \\ 
&\mathcal{N}^{(l)}_i(tir^{(l)}_i, rgb^{(l)}_i), & i=3,
\end{array}
\right.
\end{equation}
where $\mathcal{N}^{(l)}_i$ represents the fusion function of the $i$-th unit in the $l$-th layer, $rgb^{(l)}_i$ and $tir^{(l)}_i$ denote the input of the $i$-th unit in the $l$-th layer and $\mathit{O}_i$ denotes the output features of corresponding unit. Two modality features are executed separately in the 0-th and 1-th units, and different cross-modality fusion operations are performed in the 2-th and 3-th units depending on the modality input order. It is important to emphasize that units of the same type in different layers (\emph{e.g.,} $\mathcal{N}^{(l)}_i$ and $\mathcal{N}^{(l+1)}_i$) perform the same operations without sharing parameters.

\noindent\textbf{Spatial Enhancement Unit.} In simple scenes, humans can capture key objects at a glance, which indicates that single-modality information in simple scenes is sufficient for visual tasks. Moreover, in scenarios where the quality of the modality is heavily unbalanced, the interaction between the modalities brings the possibility of feature contamination. Therefore, we argue that modality interaction is not always necessary, especially for simple scenarios or scenarios with very poor quality of one modality. Inspired by this, we design an intra-modality spatial enhancement unit, which uses the context information of the modality itself to enhance the feature representation of discriminative regions.

Specifically, given a modality intermediate feature $\mathit{F} \in \mathcal{R}^{C \times H \times W}$ with $C$ channels and feature maps of size $H \times W$. We first use the spatial averaging pooling method to obtain the global statistical characteristics of the input features, denoted as $\mathit{F}_g$, and then divide $\mathit{F}$ into $G$ groups along the channel dimension, \emph{i.e.} $\mathit{F}_s, s=1,... , C/G$. Next, we measure the similarity between the global statistical feature and the local feature by a simple dot product to obtain an initial attention map. We then normalize the initial attention map and learn two scaling offset parameters ($\gamma$ and $\beta$) for each feature group to ensure the normalization can be restored. Finally, the final spatial attention map is obtained by $\mathit{Sigmoid}$ function, and the features of each location in the original feature group are enhanced. The above operations can be formulated as follows:
\renewcommand{\arraystretch}{1.5}
\begin{equation}
\begin{array}{lll}
Att_s &= \mathit{Sigmoid}(\gamma (\mathit{F}_g \cdot \mathit{F}_s) + \beta ) \\
\mathit{F_{SE}} &= \mathit{Concat}[Att_s \cdot \mathit{F}_s]  \\
s &= 1,...,C/G 
\end{array}
\end{equation}
where $\mathit{Concat}(\cdot)$ indicates the concatenation operation across the feature channel dimension. $\mathit{F_{SE}}$ represents the spatial enhanced modality feature. $G$ follows ~\cite{li2019spatial} and is set to 8 in this study.

\noindent\textbf{Channel Enhancement Unit.} Similar to SEU, we perform self-enhancement of input modal features from a channel perspective. To avoid introducing too many parameters, an efficient channel attention architecture \cite{wang2020eca} is used to calculate the attention weights. Specifically, given a modality intermediate feature $\mathit{F}$, we first use spatial pooling to obtain the aggregated feature $F_g \in \mathcal{R}^C$. Then, the weight of the channel $F^i$ is calculated only by considering the interaction between $F_i$ and its $K$ neighbors, \emph{i.e.}, which can be expressed as follows:
\renewcommand{\arraystretch}{1.5}
\begin{equation}
\begin{array}{lll}
Att_c &= \mathit{Sigmoid} \left(\sum_{k=1}^{K} \omega_{k} \cdot F^i_g \right), \quad F^i_g \in \Omega ^k_i, \\
\mathit{F_{CE}} &= Att_c \cdot \mathit{F}
\end{array}
\end{equation}
where $\Omega^k_i$ indicates the set of $K$ adjacent channels of $F^i_g$. $\omega_{k}$ denotes a fast $1D$ convolution with kernel size of $k$, and it shares parameters for all channels.  $\mathit{F_{CE}}$ represents the channel-enhanced modality feature. $k$ follows~\cite{wang2020eca} set to 3 in this study.

\noindent\textbf{RGB-to-thermal Enhancement Fusion Unit.} How to fuse RGB and thermal modality features is always a key problem in RGBT tracking. Inspired by the impressive performance of Transformer networks in various modality fusion tasks, we employ an efficient mutual attention~\cite{sun2022event} to build a cross-modal enhancement fusion unit to perform RGB-to-thermal unidirectional feature fusion. 

Unlike the self-attention block where the query (Q), key (K) and value (V) come from the same feature, this unit uses the feature of the thermal modality as the query and the feature of the RGB modality as the key and value. In particular, the two modal features of the input go through the following processing steps, including a normalization layer and a $1 \times 1$ convolutional layer with $c$ output channels to obtain the vectorized map. Subsequently, cross-modal attention is applied between the vectorized features of the two modalities. Based on this attention mechanism, the thermal modality features can be enhanced by extracting relevant features from the RGB modality, and the fusion with the thermal modality can be achieved by a simple addition operation. The entire process is expressed as follows:
\renewcommand{\arraystretch}{1.5}
\begin{equation}
\label{4}
\begin{array}{lll}
Att_{r2t} &= \mathit{Softmax}\left( \frac{Q_{tir} \cdot K_{rgb}}{\sqrt{d_k}} \right) \\
F_{r2t} &=  F_{tir} + Att_{r2t} \cdot V_{rgb} 
\end{array}
\end{equation}
where $F_{tir}$ and $F_{r2t}$ represent the original thermal modality feature and the enhancement fused thermal modality feature, respectively. In addition, the parameter $c$ is chosen to be much smaller than $H \cdot W$, where $H$ and $W$ denote the height and width of the input feature maps. This design effectively reduces the complexity of the attention operation.

\noindent\textbf{Thermal-to-RGB Enhancement Fusion Unit.} Expanding the modality fusion paradigm in another direction, we introduce a thermal-to-RGB enhancement fusion unit. Similar to RGB-to-thermal fusion, this unit employs RGB modality as query ($Q_{rgb}$) and thermal modality as key ($K_{tir}$) and value ($V_{tir}$). Then we refer to Eq \ref{4} to illustrate the process:
\renewcommand{\arraystretch}{1.5}
{\begin{equation}
\label{5}
\begin{array}{lll}
Att_{t2r} &= \mathit{Softmax}\left( \frac{Q_{rgb} \cdot K_{tir}}{\sqrt{d_k}} \right) \\
F_{t2r} &= F_{rgb} + Att_{t2r}  \cdot V_{tir}
\end{array}
\end{equation}
where $F_{rgb}$ and $F_{t2r}$ represent the original RGB modality features and enhancement fused RGB modality features, respectively. In addition, CMEU$_{r2t}$ and CMEU$_{t2r}$ are symmetric units, and bidirectional information exchange between RGB and thermal modalities can be achieved after cooperative combination.

\noindent\textbf{Router.} Each fusion unit contains a router that indicates whether the fusion unit are combined with other units.
In particular, the router is implemented by averaging pooling, a multi-layer perceptron, and two activation functions. In addition, to ensure the robustness of the decision, the router receives the information of the two modalities to make a collaborative decision. Formally, the operation of the router of $i$-unit of $l$-layer can be expressed as follows:
\renewcommand{\arraystretch}{1.5}
{\begin{equation}
\label{6}
\begin{array}{lll}
\mathcal{R}^{(l)}_i(f) &= \mathit{ReLU}\{\mathit{Tanh}[\mathit{MLP}(f_R)]\} \\
f_R &= \mathit{Concat}[\mathit{GAP}(f),\mathit{GMP}(f)] \\
f &= \mathit{Concat}[rgb^{(l)}_i,tir^{(l)}_i]
\end{array}
\end{equation}
where $\mathit{GAP}$ and $\mathit{GMP}$ respectively represent global average pooling and global maximum pooling operations. Importantly, we consider a soft version that generates continuous values as unit combination probabilities, making gradient direct propagation available.

\subsection{Dynamic Fusion Structure Space}
In this section, we analyze and discuss the dynamic fusion structure space based on a hierarchical attention network. Different from traditional fusion methods, this space can flexibly adapt to different tracking scenarios by selecting different fusion structures. Specifically, the fusion structure space consists of 4 different fusion units combined in parallel and stacked in three layers to enrich the fusion structure diversity. In addition, each unit establishes dense connections with units in other layers, making it possible for each unit to receive information from all units in the previous layer. Formally, the input to each fusion unit in this space can be expressed as follows:
\renewcommand{\arraystretch}{2}
{\begin{equation}
\label{7}
\begin{array}{lll}
O_i^{(l)} = \begin{cases} 
\sum_{j=0}^{N-1} R^{(l-1)}_{j,i} \cdot O^{(l-1)}_j & \text{if } l > 0, \\
(F_{rgb},F_{tir}) & \text{if } l = 0.
\end{cases}
\end{array}
\end{equation}
where $(F_{rgb},F_{tir})$ represent the RGB and thermal modality features extracted by the backbone network respectively. $N$ indicates the number of fusion units in each layer, and $O^{(l-1)}_j$ denotes the output of the $j$-th fusion unit in the $l\mathit{-}1$ layer. $R^{(l-1)}_{j,i} \in [0,1]$ represents the probability that the output feature from the $j$-th fusion unit of layer $l\mathit{-}1$ is sent to the $i$-th fusion unit of layer $l$. 

The selection of the fusion structure involves the modal quality difference, the tracking scene complexity, and the relationship between the target and its surrounding environment.  For example, in the case of significant quality differences between modalities, choosing a unidirectional fusion structure can minimize the influence of low-quality modalities. In contrast, the bidirectional fusion structure proved to be more effective when the two modalities exhibited complementary information.  However, due to the complexity and dynamic nature of the tracking scenario, the above examples tend to appear mixed. Therefore, the cooperation of multiple fusion structures is the key to deal with robust fusion in complex scenes.

\subsection{Discussion}
Attention-based fusion models are widely used in RGBT tracking. They may have two main limitations. First, these models often adopt several attention schemes, and the adopted schemes and their simple combination might be sub-optimal in challenging scenarios. Second, these fusion structures are fixed and thus difficult to handle various challenges over time. To handle these issues, we propose a novel attention-based fusion router (AFter) with the following three advantages. First, it is \textbf{generic}. Existing attention-based fusion methods could be regarded as a special case of AFter. Second, it is \textbf{dynamic}. AFter can automatically optimize fusion structures to handle various challenges over time. Finally, its performance is \textbf{outstanding}. It achieves outstanding performance on popular RGBT tracking datasets. Note that we have provided the evidence in experiments to demonstrate \textbf{the necessity of the dynamic fusion scheme in RGBT tracking}, as shown in rows 2 and 4 of Table~\ref{ablation_tab1}. If we statically stack these attention-based fusion units, the PR/SR scores only achieve 86.0\%/63.3\% and 68.7\%/54.1\% on RGBT234 and LasHeR datasets (Our AFter achieves 90.1\%/66.7\% and 70.3\%/55.1\%).


\section{Experiments}

\subsection{Implementation Details} 
We take ToMP~\cite{tomp} as our base tracker, which uses the first four convolution blocks of ResNet50 as feature extractors. To initialize the feature extractor parameters, we adopt the pre-trained model provided by ToMP50~\cite{tomp}, while the remaining network parameters are randomly initialized. For each sequence in a given training set, we collect the training sample and subject it to standard data augmentation operations, including rotation, translation, and scaling, aligning with the data processing scheme of the base tracker. During training, the entire model utilizes stochastic gradient descent (SGD) to minimize classification and regression loss functions. We use the LasHeR training set to train the entire tracking network in an end-to-end manner, which is used to evaluate GTOT~\cite{li2016gtot}, RGBT210~\cite{Li17rgbt210}, RGBT234~\cite{li2019rgb234}, and the LasHeR testing set ~\cite{li2021lasher}. For the evaluation of VTUAV~\cite{Zhang_CVPR22_VTUAV}, we utilize the training set from VTUAV as the training data. In addition, we assign the backbone network, IoU predictor, and model predictor a learning rate one-tenth of ToMP's default~\cite{tomp}. All parameters in the HAN are set with a learning rate of 2e-6 and a fixed iteration number of 50. AFter's implementation is conducted on the PyTorch platform and runs on a single Nvidia RTX4090 GPU with 24GB memory.


\subsection{Evaluation Dataset and Protocol}


\noindent{\bf Dataset.} To comprehensively evaluate the performance, we compare our method with previous state-of-the-art RGBT trackers on five widely used datasets.
\begin{itemize}
\item \textbf{GTOT}~\cite{li2016gtot}: This dataset serves as the first standard RGBT tracking dataset and consists of 50 RGBT video sequences registered in time and space under different scenes and conditions. It contains about 15.6K frames, and each frame is annotated with a compact bounding box. 
%
%
\item \textbf{RGBT210}~\cite{Li17rgbt210}: This dataset releases 210 pairs of RGBT video sequences totaling about 209.4K frames. According to different challenge attributes, the entire dataset is divided into 12 subsets, which are used to analyze the sensitivity of the RGBT tracker in different challenge attributes.
\item \textbf{RGBT234}~\cite{li2019rgb234}: This dataset is upgraded from RGBT210, which contains 234 pairs of aligned RGBT videos totaling about 233.4K frames. In addition, it also considers 12 challenge attributes, which could more comprehensively evaluate the robustness of the RGBT trackers under various challenge attributes.
\item \textbf{LasHeR}~\cite{li2021lasher}: This dataset is the largest RGBT tracking dataset, containing 1224 aligned video sequences and up to 1469.6K frames. Besides, it also divides the training set and the testing set, which contains 979 pairs of RGBT video and 245 pairs of RGBT video respectively, providing a platform for fair comparison of different algorithms.
\item \textbf{VTUAV}~\cite{Zhang_CVPR22_VTUAV}: This dataset collects RGBT data from UAV scenarios, broadening the practice of RGBT tracking. It contains 500 aligned RGBT video sequences and up to 1.7 million frames. We primarily focus our experiments on its short-term tracking subset. 
%
\end{itemize}


\noindent{\bf Protocol.} In RGBT tracking typically adopts the Precision Rate (PR), Success Rate (SR) and Normalised Precision Rate (NPR) from One-Pass Evaluation (OPE) as metrics for quantitative performance evaluation, which are widely used in current RGBT tracking studies. 
%
\begin{itemize}
\item \textbf{PR}: the PR evaluates the percentage of total frames where the Euclidean distance between the centroid of the tracking result and the centroid of the truth enclosing frame is less than the threshold $\tau$, and is primarily a measure of the tracker's ability to locate the target in terms of location. In the GTOT dataset, $\tau$ was set to five pixels due to the small target size, and $\tau$ was set to 20 pixels for the other datasets to obtain a representative PR score.
\item \textbf{SR}: the SR evaluates the percentage of total frames for which the intersection ratio between the tracking result and the truth-enveloped frame is greater than the threshold $\delta$, and primarily measures the ability of the tracker in terms of scale estimation. By varying $\delta$ to construct success curves, tracking performance can be demonstrated for different levels of intersection and merger ratios. Unlike the representative PR score, which is obtained using a fixed threshold, the representative SR score is obtained by calculating the area under the success rate curve.
\item \textbf{NPR}: image resolution and target size have a significant impact on the calculation of PR. In order to eliminate these effects, the PR is normalised using the scale of the truth-envelope box when calculating the PR. The normalised accuracy curve can be obtained by changing the normalisation threshold, and the area under the normalised accuracy curve with the normalisation threshold in the range $[0,0.5]$ is calculated as the representative NPR score.
\end{itemize}

\subsection{Quantitative Comparison}
We evaluate our algorithm on five popular RGBT tracking benchmarks and compare its performance with current state-of-the-art trackers. The effectiveness of our proposed method is demonstrated in Table~\ref{overall_result_tab1}, which provides a summary of the comparison results.

\begin{table*}[ht]
	\centering
	\caption{The PR, NPR, and SR scores (\%) of various trackers on five datasets. The best and second results are in $\textcolor{red} red$ and $\textcolor{blue} blue$ colors, respectively. }
        \renewcommand\arraystretch{1.4}{
	\resizebox{1\textwidth}{!}{
	\begin{tabular}{c|c|c|cc|cc|cc|ccc|cc|c}
		\hline
		\multirow{2}{*}{Methods} & \multirow{2}{*}{Pub. Info.} & \multirow{2}{*}{Backbone} & \multicolumn{2}{c|}{GTOT} & \multicolumn{2}{c|}{RGBT210} & \multicolumn{2}{c|}{RGBT234} & \multicolumn{3}{c|}{LasHeR} & \multicolumn{2}{c|}{VTUAV} & {FPS} \\
		& & & PR$\uparrow$ & SR$\uparrow$ & PR$\uparrow$ & SR$\uparrow$ & PR$\uparrow$ & SR$\uparrow$ & PR$\uparrow$ & NPR$\uparrow$ & SR$\uparrow$ & PR$\uparrow$ & SR$\uparrow$ & $\uparrow$ \\
		\hline
        MANet~\cite{li2019manet} & ICCVW 2019 & VGG$-$M & 89.4 & 72.4 & $-$ & $-$ & 77.7 & 53.9 & 45.5 & 38.3 & 32.6  & $-$ & $-$ & 1\\
        DAPNet~\cite{zhu2019dense} & ACM MM 2019 & VGG$-$M & 88.2 & 70.7 & $-$ & $-$ & 76.6 & 53.7 & 43.1 & 38.3 & 31.4 & $-$ & $-$ & 2\\
        mfDiMP~\cite{zhang2019multi} & ICCVW 2019 & ResNet$-$50 & 83.6 & 69.7 & 78.6 & 55.5 & $-$ & $-$ & 44.7 & 39.5 & 34.3 & 67.3 & 55.4  & 10.3 \\
        CMPP~\cite{2020CMPP} & CVPR 2020 & VGG$-$M & \textcolor{red}{92.6} & 73.8 & $-$ & $-$ & 82.3 & 57.5 & $-$ & $-$ & $-$ & $-$ & $-$  &1.3\\
        MaCNet~\cite{zhang2020MaCNet} & Sensors 2020 & VGG$-$M & 88.0 & 71.4 & $-$ & $-$ & 79.0 & 55.4 & 48.2 & 42.0 & 35.0 & $-$ & $-$  & 0.8\\
        CAT~\cite{2020CAT} & ECCV 2020 & VGG$-$M & 88.9 & 71.7 & 79.2 & 53.3 & 80.4 & 56.1 & 45.0 & 39.5 & 31.4 & $-$ & $-$  & 20\\
        FANet~\cite{2020FANet} & TIV 2021 & VGG$-$M & 89.1 & 72.8 & $-$ & $-$ & 78.7 & 55.3 & 44.1 & 38.4 & 30.9 & $-$ & $-$  & 19\\
        ADRNet~\cite{ADRNet2021} & IJCV 2021 & VGG$-$M & 90.4 & 73.9 & $-$ & $-$ & 80.7 & 57.0 & $-$ & $-$ & $-$ & 62.2 & 46.6 & 25\\
        JMMAC~\cite{2020JMMAC} & TIP 2021 & VGG$-$M & 90.2 & 73.2 & $-$ & $-$ & 79.0 & 57.3 & $-$ & $-$ & $-$ & $-$ & $-$  & 4\\
        MANet$++$~\cite{2021MANet++} & TIP 2021 & VGG$-$M & 88.2 & 70.7 & $-$ & $-$ & 80.0 & 55.4 & 46.7 & 40.4 & 31.4 & $-$ & $-$  & 25.4 \\
        APFNet~\cite{APFNet2022} & AAAI 2022 & VGG$-$M & 90.5 & 73.7 & $-$ & $-$ & 82.7 & 57.9 & 50.0 & 43.9 & 36.2 & $-$ & $-$  & 1.3\\
        DMCNet~\cite{DMCNet2022} & TNNLS 2022 & VGG$-$M & 90.9 & 73.3 & 79.7 & 55.5 & 83.9 & 59.3 & 49.0 & 43.1 & 35.5 & $-$ & $-$  &2.3 \\
        ProTrack~\cite{ProRGBTTrack} & ACM MM 2022 & ViT$-$B & $-$ & $-$ & $-$ & $-$ & 78.6 & 58.7 & 50.9 & $-$ & 42.1 & $-$ & $-$  & 30 \\
        FTNet~\cite{FT2022} & AVSS 2022 & VGG$-$M & 91.2 & 73.6 & $-$ & $-$ & 83.7 & 60.1 & 52.6 & $-$ & 38.1 & $-$ & $-$  & 1.2 \\
        MIRNet~\cite{hou2022mirnet} & ICME 2022 & VGG$-$M & 90.9 & 74.4 & $-$ & $-$ & 81.6 & 58.9 & $-$ & $-$ & $-$ & $-$ & $-$  & 30 \\
        HMFT~\cite{Zhang_CVPR22_VTUAV} & CVPR 2022 & ResNet$-$50 & 91.2 & \textcolor{blue}{74.9} & 78.6 & 53.5 & 78.8 & 56.8 & $-$ & $-$ & $-$& \textcolor{blue}{75.8} & \textcolor{blue}{62.7}   & \textcolor{blue}{30.2} \\
        MFG~\cite{wang2022mfgnet} & TMM 2022 & ResNet$-$18 & 88.9 & 70.7 & 74.9 & 46.7 & 75.8 & 51.5 & $-$ & $-$ & $-$& $-$ & $-$   & $-$ \\
        DFNet~\cite{peng2022dynamic} & TITS 2022 & VGG$-$M & 88.1 & 71.9 & $-$ & $-$ & 77.2 & 51.3 & $-$ & $-$ & $-$& $-$ & $-$   & $-$ \\
        DRGCNet~\cite{mei2023differential} & IEEE SENS J 2023 & VGG$-$M & 90.5 & 73.5 & $-$ & $-$ & 82.5 & 58.1 & 48.3 & 42.3 & 33.8 & $-$ & $-$  & 4.9 \\
        CMD~\cite{zhang2023efficient} & CVPR 2023 & ResNet$-$50 & 89.2 & 73.4 & $-$ & $-$ & 82.4 & 58.4 & 59.0 & 54.6 & 46.4 & $-$ & $-$  & 30 \\
        \hline
        ViPT~\cite{ViPT} & CVPR 2023 & ViT$-$B & $-$ & $-$ & $-$ & $-$ & 83.5 & 61.7 & 65.1 & $-$ & 52.5 & $-$ & $-$  & $-$ \\
        TBSI~\cite{TBSI} & CVPR 2023 & ViT$-$B & $-$ & $-$ & \textcolor{blue}{85.3} & \textcolor{blue}{62.5} & \textcolor{blue}{87.1} & 63.7 & \textcolor{blue}{69.2} & \textcolor{blue}{65.7} & \textcolor{red}{55.6} & $-$ & $-$  & \textcolor{red}{36.2} \\
        CAT$++$~\cite{liu2024rgbt} & TIP 2024 & VGG$-$M & 91.5 & 73.3 & 82.2 & 56.1 & 84.0 & 59.2 & 50.9 & 44.4 & 35.6 & $-$ & $-$  & 14 \\
        OneTracker~\cite{OneTracker} & CVPR 2024 & ViT$-$B & $-$ & $-$ & $-$ & $-$ & 85.7 & \textcolor{blue}{64.2} & 67.2 & $-$ & 53.8 & $-$ & $-$  & $-$ \\
        {Un$-$Track}~\cite{Un-Track} & CVPR 2024 & ViT$-$B & $-$ & $-$ & $-$ & $-$ & 84.2 & 62.5 & 66.7 & $-$ & 53.6 & $-$ & $-$  & $-$ \\
        SDSTrack~\cite{SDSTrack} & CVPR 2024 & ViT$-$B & $-$ & $-$ & $-$ & $-$ & 84.8 & 62.5 & 66.5 & $-$ & 53.1 & $-$ & $-$  & 20.86 \\
  		\hline
		AFter &  $-$ & ResNet$-$50 & \textcolor{blue}{91.6} & \textcolor{red}{78.5} & \textcolor{red}{87.6} & \textcolor{red}{63.5} & \textcolor{red}{90.1} & \textcolor{red}{66.7} & \textcolor{red}{70.3} & \textcolor{red}{65.8} & \textcolor{blue}{55.1} & \textcolor{red}{84.9} & \textcolor{red}{72.5} & 23 \\ \hline
       \end{tabular}}}
	\label{overall_result_tab1}
\end{table*}


\noindent{\bf Evaluation on GTOT dataset.} The results of the comparison on the GTOT dataset are shown in Table~\ref{overall_result_tab1}. Our method outperforms state-of-the-art trackers on the GTOT dataset, with gains over HMFT, FTNet, and MIRNet in PR/SR by 0.4\%/3.6\%, 0.4\%/4.9\%, and 0.7\%/4.1\%, respectively. We further compare our method with CMPP, the current top-performer on this dataset. Although our PR is 1.0\% lower than CMPP's, our AFter surpasses CMPP by 4.7\% in SR, indicating superior performance in target scale regression. The lower PR can be attributed to the prevalence of small objects in the GTOT dataset. CMPP's feature pyramid strategy and historical information pool improve feature representation and current frame depiction, respectively. However, these strategies considerably affect CMPP's efficiency, making our AFter approximately 18 times faster. 



\noindent{\bf Evaluation on RGBT210 dataset.} As shown in Table~\ref{overall_result_tab1}, AFter surpasses all state-of-the-art trackers on the RGBT210 dataset. Compared to mfDiMP~\cite{zhang2019multi}, the VOT2019-RGBT winner, AFter achieves substantial PR/SR improvements with gains of 9.0\%/8.0\%. Moreover, compared to TBSI~\cite{TBSI}, it is the second best performing algorithm in this dataset, our method exhibits a 2.3\%/1.0\% performance advantage in PR/SR.



\noindent{\bf Evaluation on RGBT234 dataset.} To further evaluate the effectiveness of AFter, we perform a series of experiments on the RGBT234 dataset, encompassing both overall and attribute-based comparisons.


\textit{1) Overall Comparison:} Our method is evaluated against 24 state-of-the-art RGBT trackers on the RGBT234 dataset, with results presented in Table~\ref{overall_result_tab1}. AFter outperforms all the state-of-the-art RGBT methods in PR/SR metrics, demonstrating its effectiveness. Compared to TBSI, which is the second best performing algorithm in this dataset, AFter exhibits a 2.0\%/3.0\% performance advantage on the PR/SR metric. In particular, although TBSI elaborately designs a bi-directional fusion module bridging multimodal templates, its fixed fusion structure is hard to adapt to dynamic tracking scenarios on the one hand, and the single-layer fusion space is insufficient to fully fuse modal features on the other hand. In contrast, AFter adopts a hierarchical attention network that can dynamically adapt to different tracking scenarios, and can achieve sufficient fusion of different modality features through the stacking of multi-layer fusion space. 
In addition, AFter also outperforms several latest algorithms, including SDSTrack~\cite{SDSTrack}, Un-Track~\cite{Un-Track} and OneTracker~\cite{OneTracker} by 5.3\%/4.2\% , 5.9\%/4.2\%, and 4.4\%/3.5\% in the PR/SR metrics, respectively.


\begin{figure}[ht]
	\centering
	\includegraphics[width=0.8\linewidth]{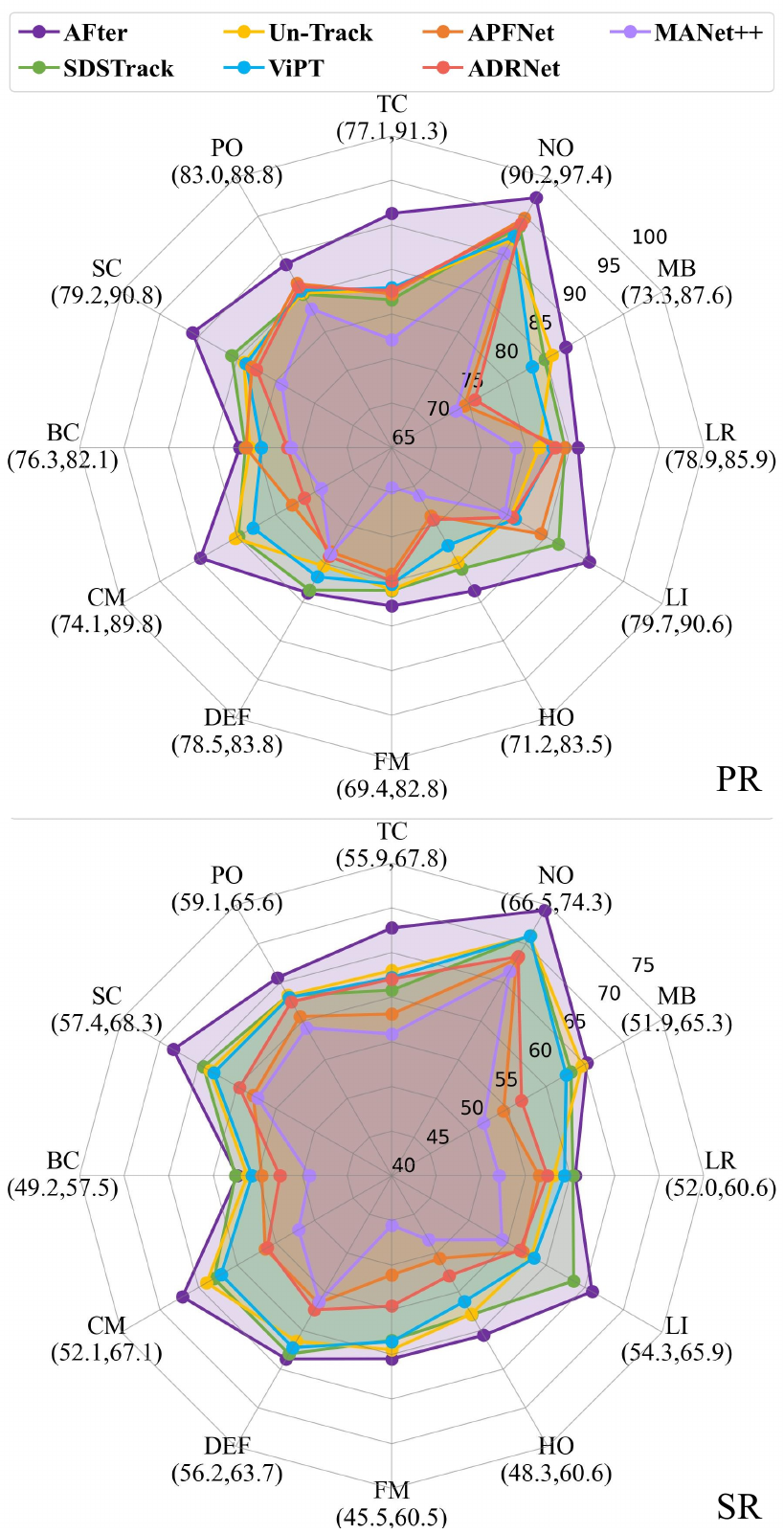}
	\caption{Precision Rate (PR) and Success Rate (SR) of challenge attributes on the RGBT234 dataset.}
	\label{fig::radar_results}
\end{figure}

\textit{2) Challenge-based Comparison:} To further verify the advantages of our approach in different challenging scenarios, we compare our AFter with other state-of-the-art RGBT trackers, including SDSTrack~\cite{SDSTrack}, Un-Track~\cite{Un-Track}, ViPT~\cite{ViPT}, APFNet~\cite{APFNet2022}, MANet++~\cite{2021MANet++}, and ADRNet~\cite{ADRNet2021} on subsets with different challenge attributes. 
The challenge attributes include no occlusion (NO), partial occlusion (PO), heavy occlusion (HO), low illumination (LI), low resolution (LR), thermal crossover (TC), deformation (DEF), fast motion (FM), scale variation (SV), motion blur (MB), camera moving (CM), and background clutter (BC). The evaluation results are shown in Fig.~\ref{fig::radar_results}, where the marks of each corner represent the attributes of the challenge subset and the highest and lowest performance under that attribute, respectively.

Specifically, AFter significantly outperforms existing state-of-the-art algorithms on the challenge subsets of SC, CM, LI, and TC, with improvements of 11.6\%/10.9\%, 15.7\%/15.0\%, 10.9\%/11.6\%, and 13.4\%/11.9\% in the PR/SR metrics, respectively. We attribute this mainly to the fact that these challenge subsets consist of low-quality bimodal or extremely low-quality unimodal tracking scenarios, which pose significant challenges for the conventional bidirectional fusion strategies commonly used in existing methods. In contrast, AFter provides more flexible fusion strategies, including self-fusion, unidirectional fusion, and bidirectional fusion, and thus can provide better fusion results for the above scenarios.
The experiment indicates that the superior performance of AFter under a variety of challenging scenarios, demonstrating the robustness of AFter in handling adverse conditions. 
%


\noindent{\bf Evaluation on LasHeR dataset.} To comprehensively evaluate the effectiveness of the proposed AFter method against state-of-the-art trackers, we conduct experiments on the LasHeR testing set with 15 RGBT trackers. The evaluation results are shown in Table~\ref{overall_result_tab1}. AFter significantly outperforms almost all existing trackers on the LasHeR testing set. Although our method is 0.5\% lower than TBSI in the SR metric, it exceeds TBSI by 1.1\%/0.1\% in the PR/NPR metric. The minor performance disadvantage on the SR metrics we could attribute to the longer and more challenging sequences in the LasHeR dataset, as well as the fact that TBSI leverages multi-modal template features to guide the fusion, , which is more advantageous in long sequences due to the introduction of historical information. In the future, AFter will also consider introducing template features or historical time series information to further enhance the fusion effect.
In addition, AFter outperforms SDSTrack, ViPT and CMD by 3.8\%/-\%/2.0\%, 5.2\%/-\%/2.6\% and 11.3\%/11.2\%/8.7\% in PR/NPR/SR metrics, respectively. These results demonstrate the effectiveness of our method.

\noindent{\bf Evaluation on VTUAV dataset.} We evaluate the performance of our AFter method on the recently released VTUAV drone perspective RGBT tracking dataset. The results, presented in Table~\ref{overall_result_tab1}, demonstrate the remarkable performance of AFter, thus confirming its effectiveness. Specifically, AFter achieves an impressive PR score of 84.9\% and a commendable SR score of 72.5\%. These scores indicate that AFter still exhibits high accuracy and robustness in RGBT tracking from the UAV perspective. Specifically, AFter outperforms the current leading HMFT method by a significant margin, surpassing it by 9.1\% and 9.8\% in terms of PR and SR, respectively. These results further solidify its role as a state-of-the-art method in RGBT tracking tasks.


\subsection{Ablation Study}
In this section, we conduct several ablation studies on the RGBT234 dataset and LasHeR testing set to confirm the effectiveness of our proposed method.

\begin{table}[]
\centering
\setlength{\tabcolsep}{2mm}{
\caption{Ablation studies of fusion layers and routers in HAN.}
\label{ablation_tab1}
\renewcommand{\arraystretch}{1.9}
\resizebox{0.44\textwidth}{!}{
\begin{tabular}{cccccccc}
\toprule
\multirow{2}{*}{\textbf{Layers}} & \multirow{2}{*}{\textbf{Router}} & \multicolumn{2}{c}{\textbf{RGBT234}} & \multicolumn{2}{c}{\textbf{LasHeR}} & \multicolumn{2}{c}{\textbf{HAN}} \\ 
\cmidrule(lr){3-4} 
\cmidrule(lr){5-6} 
\cmidrule(lr){7-8} 
   &  & \textbf{PR}    & \textbf{SR}             & \textbf{PR}     & \textbf{SR}    & \textbf{Params} & \textbf{FLOPs}        \\ \hline
\textbf{2}                       & \checkmark                       & 88.9 & 65.4         & 68.2   & 53.6 & 5.26M & 0.35G        \\ 
\rowcolor{cyan!10} \textbf{3}                       & \checkmark                       & 90.1 & 66.7         & 70.3   & 55.1 & 7.89M & 0.52G        \\ 
\textbf{4}                       & \checkmark                       & 88.2 & 64.3         & 68.5   & 53.7 & 10.52M & 0.69G        \\ \hline
\textbf{1}                       & ×                                & 86.3 & 63.1         & 67.2   & 52.9 & 0.52M & 0.17G        \\ 
\textbf{3}                       & ×                                & 86.0 & 63.3         & 68.7   & 54.1 & 1.57M & 0.51G        \\ 
\bottomrule
\end{tabular}}}
\end{table}

\begin{table}[]
\centering
\setlength{\tabcolsep}{2mm}{
\caption{Ablation studies of fusion units in HAN.}
\label{ablation_tab2}
\renewcommand{\arraystretch}{1.9}
\resizebox{0.48\textwidth}{!}{
\begin{tabular}{cccccccccc}
\toprule
\multirow{2}{*}{\textbf{N0}} & \multirow{2}{*}{\textbf{N1}} & \multirow{2}{*}{\textbf{N2}} & \multirow{2}{*}{\textbf{N3}} & \multicolumn{2}{c}{\textbf{RGBT234}} & \multicolumn{2}{c}{\textbf{LasHeR}} & \multicolumn{2}{c}{\textbf{HAN}} \\
\cmidrule(lr){5-6} 
\cmidrule(lr){7-8} 
\cmidrule(lr){9-10} 
   & & & & \textbf{PR}    & \textbf{SR}             & \textbf{PR}     & \textbf{SR}    & \textbf{Params} & \textbf{FLOPs}        \\ \hline
×                   & \checkmark                   & \checkmark                   & \checkmark                   & 87.3         & 65.1         & 67.7         & 53.2        & 6.31M           & 0.51G         \\ \hline
\checkmark                   & ×                   & \checkmark                   & \checkmark                   & 88.9         & 65.7         & 67.5         & 53.3        & 6.32M           & 0.52G         \\ \hline
\checkmark                   & \checkmark                   & ×                   & \checkmark                   & 87.5         & 64.4         & 67.8         & 53.6        & 5.52M           & 0.26G         \\ \hline
\checkmark                   & \checkmark                   & \checkmark                   & ×                   & 88.6         & 65.6         & 67.9         & 53.3        & 5.52M           & 0.26G         \\ \hline
\rowcolor{cyan!10} \checkmark                   & \checkmark                   & \checkmark                   & \checkmark                   & 90.1         & 63.3         & 68.7         & 54.1        & 7.89M           & 0.52G         \\ \bottomrule
\end{tabular}}}
\end{table}

\noindent{\bf Analysis of HAN.} We evaluate HAN with different layers and summarize the results in Table \ref{ablation_tab1}. It can be observed that when the number of layers is 3, HAN achieves optimal performance in both data sets, which demonstrates the reasonableness of the HAN layer setting. To study the effectiveness of the router, we remove it and the HAN degrades to a dense fixed fusion module. As shown in the last two rows of table \ref{ablation_tab1}, it can be observed that removing the routers results in a significant performance degradation regardless of whether the number of layers is set to 1 or 3. This experiment shows that simply aggregating multiple different attentions is not enough and validates the importance of dynamic fusion structures. 
In addition, we show the parameter quantities and computational quantities for all variants in Table \ref{ablation_tab1}. It can be observed that HAN as a whole is lightweight and its main parametric quantities are derived from the router.


\begin{table}[]
\centering
\setlength{\tabcolsep}{2mm}{
\caption{Ablation studies of HAN on different frameworks}
\label{ablation_tab2}
\renewcommand{\arraystretch}{1.9}
\resizebox{0.44\textwidth}{!}{
\begin{tabular}{cccccc}
\toprule
\multirow{2}{*}{\textbf{Method}} & \multicolumn{2}{c}{\textbf{RGBT234}} & \multicolumn{3}{c}{\textbf{LasHeR}} \\ 
\cmidrule(lr){2-3} 
\cmidrule(lr){4-6}
                        & \textbf{PR}           & \textbf{SR}           & \textbf{PR}      & \textbf{NPR}     & \textbf{PR}     \\ \hline
\textbf{DiMP~\cite{DiMP}$_{(RGBT)}$}                    & 79.2         & 56.5         & 52.9    & 47.9    & 39.2   \\ 
\textbf{OSTrack~\cite{ostrack}$_{(RGBT)}$}                 & 86.7         & 64.8         & 66.5    & 63.0    & 53.0   \\ 
\textbf{ToMP~\cite{tomp}$_{(RGBT)}$}                    & 86.8         & 65.1         & 65.4    & 61.1    & 51.6   \\ \hline
\textbf{DiMP$_{(RGBT)}$ +  HAN}                & 88.2         & 62.2         & 62.4    & 58.6    & 49.3   \\ 
\textbf{OSTrack$_{(RGBT)}$ +  HAN}             & 87.7         & 65.7         & 68.1    & 64.5    & 54.2   \\ 
\rowcolor{cyan!10} \textbf{ToMP$_{(RGBT)}$ +  HAN}                & 90.1        & 66.7        & 70.3    & 65.8   & 55.1   \\ 
\bottomrule
\end{tabular}}}
\end{table}

\noindent{\bf Generalization of HAN.} To verify the generalization of HAN, we first extend three different RGB tracking frameworks (DiMP~\cite{DiMP}, OSTrack~\cite{ostrack}, ToMP~\cite{tomp}) into RGBT trackers using a late fusion approach. We then embed the HAN into these trackers and evaluate their performance on the RGBT234 dataset and the LasHeR test set. Table~\ref{ablation_tab2} shows that these trackers effectively improve their performance after incorporating HAN, which demonstrates the generality of HAN.  In particular, ToMP-HAN outperforms ToMP$_{RGBT}$ by achieving a 3.3\%/1.6\% increase in the PR/SR metric on RGBT234 and a 4.9\%/4.7\%/3.5\% increase on the LasHeR testing set. These results confirm the effectiveness of HAN.

%

\begin{figure*}[ht]
    \centering
    \includegraphics[width=0.9\linewidth]{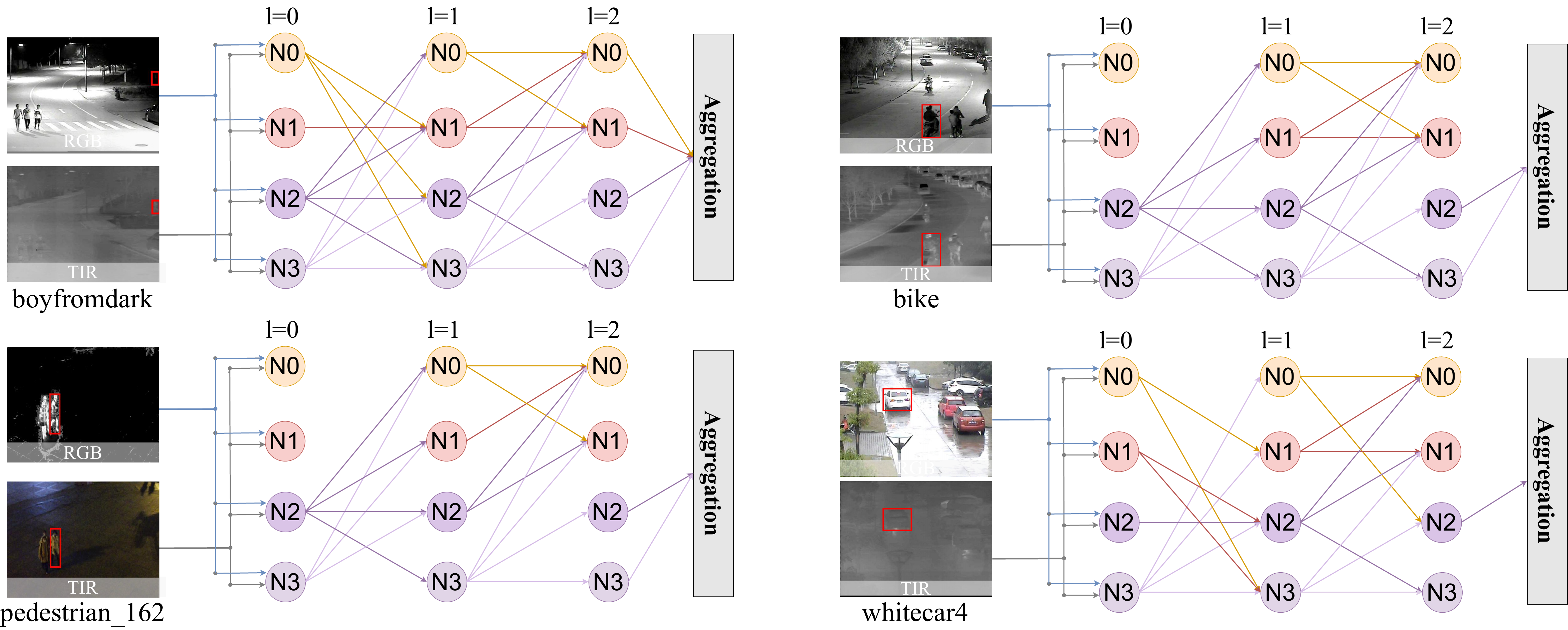}
    \caption{ Visualization of fusion structures in HAN under different scenarios.}
    \label{dfm_visual}
\end{figure*}

\begin{figure*}[ht]
    \centering
    \includegraphics[width=0.9\linewidth]{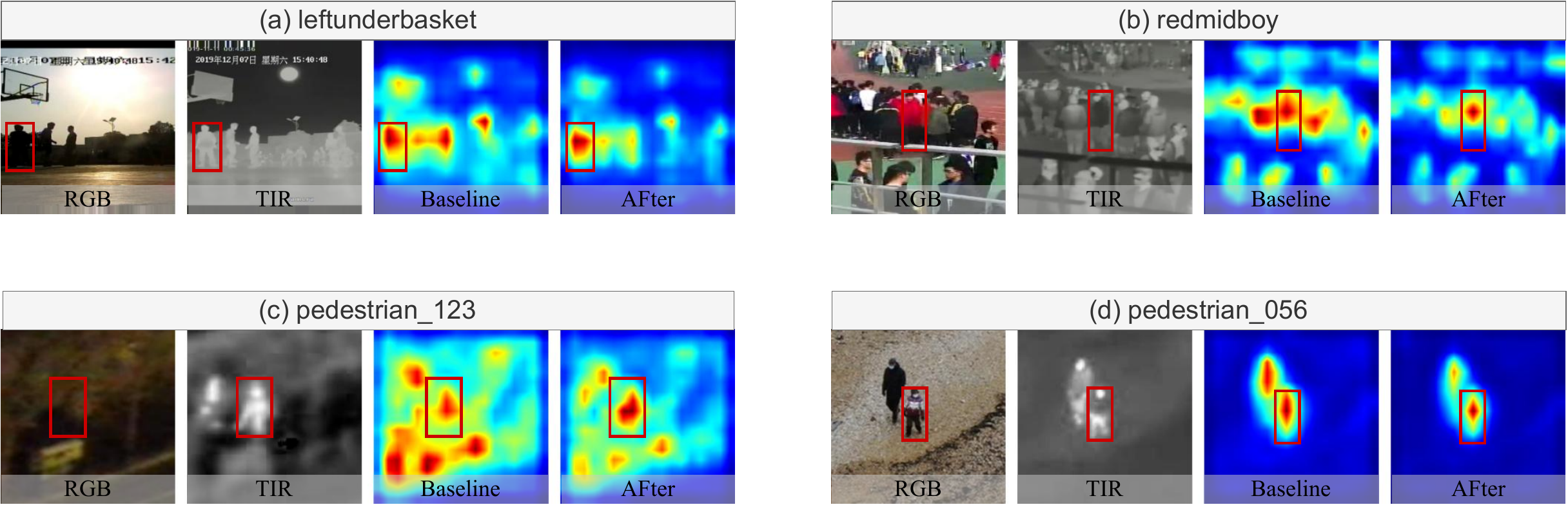}
    \caption{Comparison of feature maps between ToMP$_{RGBT}$ (third column) and HAN-based ToMP$_{RGBT}$ (fourth column).}
    \label{dfm_feats}
\end{figure*}

\noindent{\bf Dynamic structure visualization of HAN.} To observe the dynamic fusion structure in the HAN, we visualize the structure of the HAN under several representative sample inputs in Figure \ref{dfm_visual}. It can be seen that in complex input scenarios, the HAN connects more fusion units, as shown in the $boyfromdark$ sequence. In contrast, in relatively simple input scenarios, such as the $pedestrian\_162$ sequence, partial connections between fusion units are established in the HAN. A similar phenomenon occurs in the other two sequences. The experiment demonstrates that HAN can dynamically combine different quantities of fusion units based on the complexity of the input scenario, achieving a frame-level dynamic fusion process.

\noindent{\bf Fusion feature visualization of HAN.} To further validate the effectiveness of HAN, we visualize the fusion feature maps before and after applying HAN in Fig.~\ref{dfm_feats}. It can be observed that baseline fusion strategies struggle with accurate target location in complex scenarios, while our approach performs tracking robustly and accurately. 
For instance, in Fig.~\ref{dfm_feats} (b), although the color information of the target is strongly discriminative in RGB modality and more hard to discriminate in TIR modality. Therefore, the baseline fusion scheme leads to inaccurate multi-peak responses due to its difficulty in collaborating modality information of different qualities.
In Fig.~\ref{dfm_feats} (c), both modalities suffer from the challenge of low resolution due to the small size of the pedestrians in the UAV view. However, the fixed structure of the baseline fusion strategy is poor to handle the effective fusion of the low quality features of the two modalities, which again brings about the multi-peak response issue.
In contrast, benefiting from the dynamic fusion structure of the HAN, for different tracking scenarios, it always presents a single-peak response for accurately location the target. Similar phenomena are also occurring in Fig.~\ref{dfm_feats} (a) and (d).
Hence, HAN can provide effective fusion modes and potentially contribute valuable insights for future fusion pattern design.
%

\begin{figure*}[ht]
    \centering
    \includegraphics[width=1\linewidth]{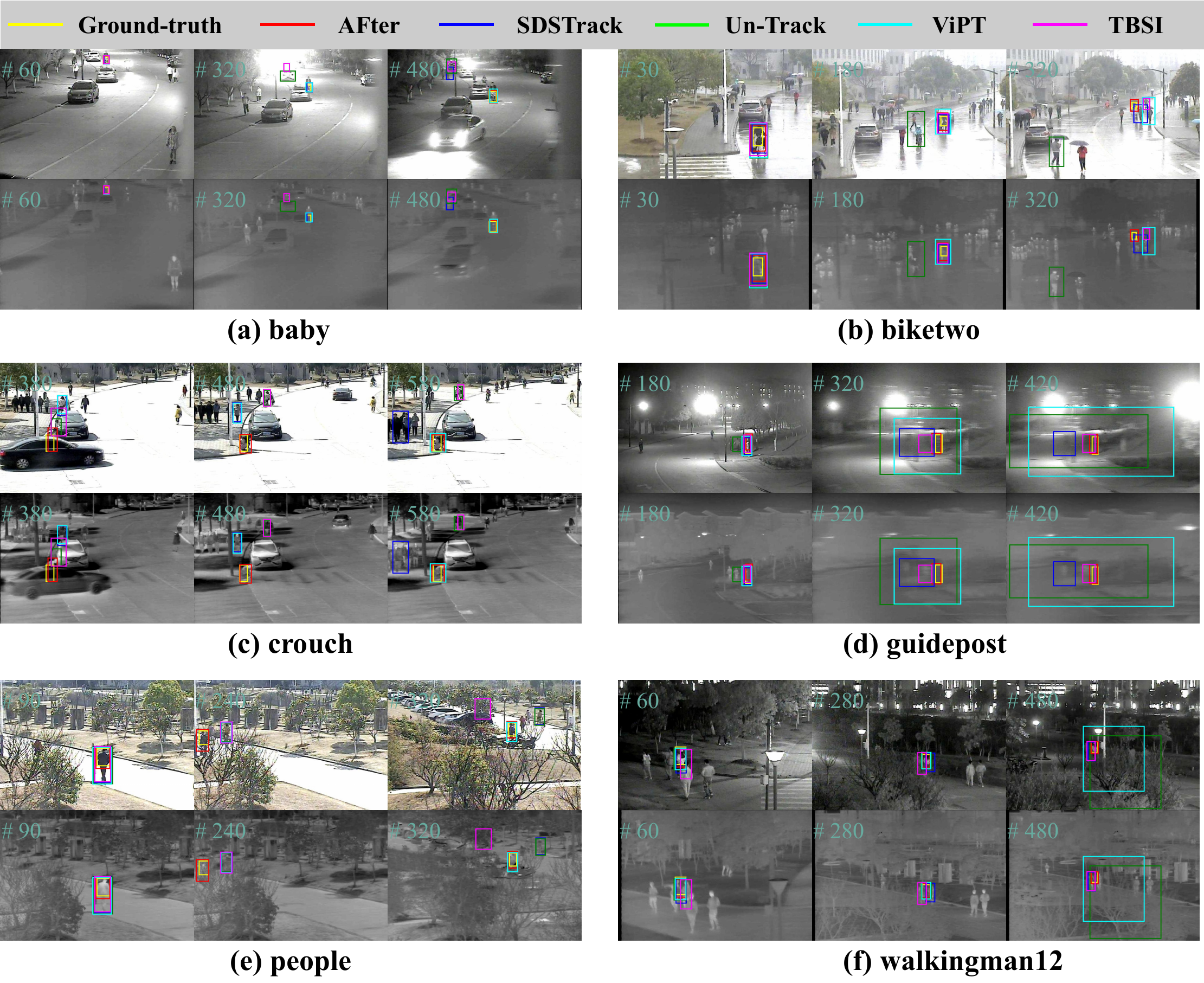}
    \caption{Qualitative comparison of AFter against four state-of-the-art trackers on six video sequences.}
    \label{success}
\end{figure*}

\noindent{\bf Visualization of tracking results.}
To visually demonstrate the effectiveness of the proposed AFter, we take some advanced trackers to compare in Fig.~\ref{success}, including SDSTrack~\cite{SDSTrack}, Un-Track~\cite{Un-Track}, ViPT~\cite{ViPT} and TBSI~\cite{TBSI}. 
In particular, we present the visual tracking results of the above tracker on six representative sequences, all from the RGBT234 dataset. For clarity, we provide three frame pairs for each sequence. It can be seen that our method clearly outperforms the other trackers in several challenges such as low resolution, low illumination, severe occlusion, thermal crossings, and fast motion.
For instance, the $baby$ sequence in Fig~\ref{success} (a) is very challenging in a low-resolution and low-illumination scene, and most trackers lose the target after a long period of tracking, while our approach can consistently locate the target. The person in Fig~\ref{success} (e) is in a fast-moving and heavily occluded scene, only AFter is able to locate the target robustly at frame 240, while all other algorithms fail to track when the person is occluded.
%

\begin{figure}[ht]
    \centering
    \includegraphics[width=0.9\linewidth]{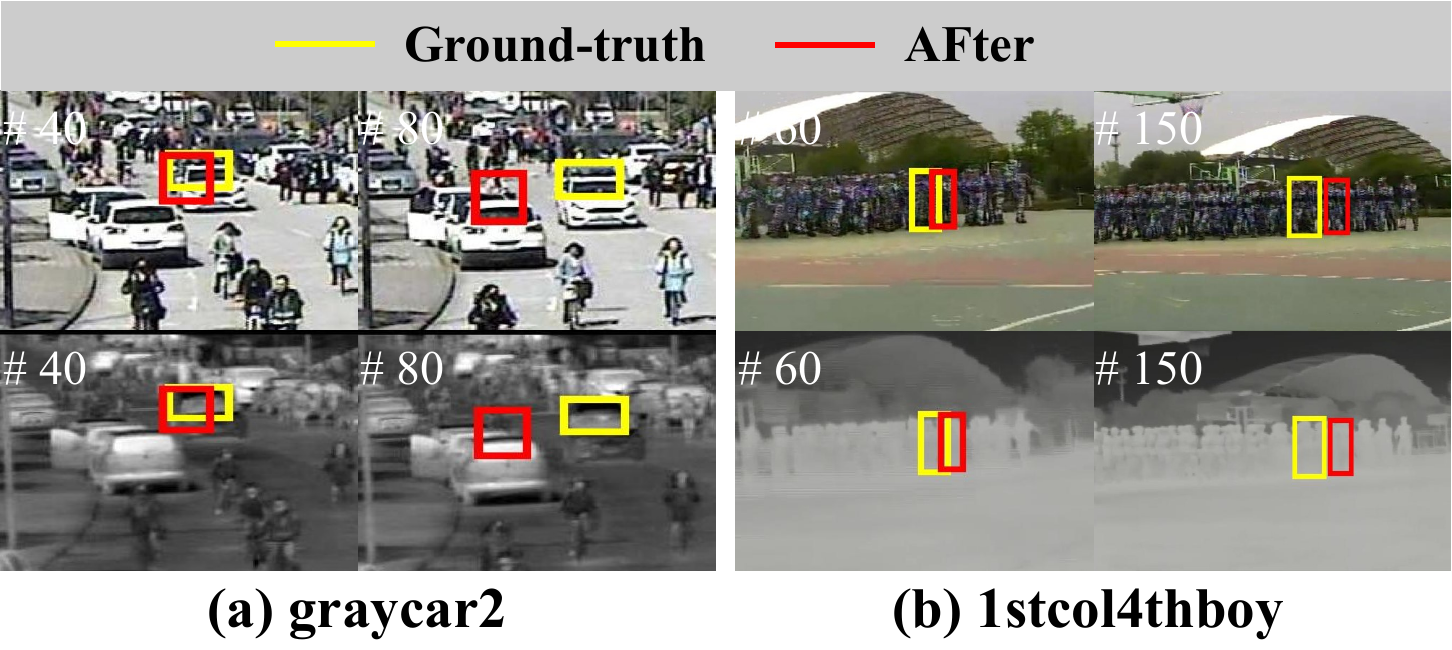}
    \caption{two instances of AFter tracking failure.}
    \label{failed}
\end{figure}

In Fig.~\ref{failed}, we also show two instances of AFter tracking failure. In (a), the severe occlusion of the target causes a significant loss of information from both modalities, resulting in tracking failure. In (b), the tracking failure is caused by the interference from similar objects, which occurs simultaneously in both modalities. In the future, we will investigate spatial-temporal models in the framework to address the challenges of severe occlusion and similar appearance.



%

%

\section{Conclusion}
In this work, we present a novel attention-based fusion router (AFter) for RGBT tracking, which is the first investigation of dynamic multi-modal feature fusion in the field of RGBT tracking. We also design a hierarchical attention network to construct a scene-aware dynamic fusion structure space, which overcomes the limitation of the fixed fusion structure in the existing methods. The superior performance of AFter over existing RGBT trackers is demonstrated across five mainstream RGBT tracking datasets, highlighting its robustness and effectiveness. Noting that although embedding routers in each unit can reduce the decision difficulty, they also bring a certain computational burden. In the future, we plan to improve the efficiency of the fusion structure selection mechanism and explore more effective ways to navigate the space of the fusion structure.







\bibliographystyle{IEEEtran}
\bibliography{After}

\end{document}